\documentclass[10pt,twocolumn,letterpaper]{article}

\usepackage[pagenumbers]{cvpr} 

\usepackage{amsmath,amsfonts,bm}
\usepackage{colortbl}
\usepackage[dvipsnames]{xcolor}

\usepackage{multirow, hhline}

\definecolor{cvprblue}{rgb}{0.21,0.49,0.74}
\usepackage[pagebackref,breaklinks,colorlinks,citecolor=cvprblue]{hyperref}
\usepackage{algorithm}
\usepackage{algorithmic}
\usepackage{balance}
\def\paperID{xxxx} 
\def\confName{CVPR}
\def\confYear{2024}

\usepackage{multirow}

\title{Seeing and Hearing: Open-domain Visual-Audio Generation with Diffusion Latent Aligners} 
\author{Yazhou Xing$^{1}$\thanks{equal contribution}  \quad Yingqing He$^{1 *}$ \quad Zeyue Tian$^{1 *}$ \quad Xintao Wang$^{2}$ \quad Qifeng Chen$^{1}$ 
\and
$^1$HKUST \quad $^2$ARC Lab, Tencent PCG
}

\begin{document}
\newcommand{\revise}[1]{\textcolor{red}{#1}}
\newcommand{\tocite}{\textcolor{red}{TO Cite}}

\newcommand{\var}[1]{\text{\texttt{#1}}}
\newcommand{\func}[1]{\text{\textsl{#1}}}
\makeatletter
\newcounter{phase}[algorithm]
\newlength{\phaserulewidth}
\newcommand{\setphaserulewidth}{\setlength{\phaserulewidth}}
\newcommand{\phase}[1]{%
  \vspace{-1.25ex}
  \Statex\leavevmode\llap{\rule{\dimexpr\labelwidth+\labelsep}{\phaserulewidth}}\rule{\linewidth}{\phaserulewidth}
  \Statex\strut\refstepcounter{phase}\textit{Stage~\thephase~--~#1}
  \vspace{-1.25ex}\Statex\leavevmode\llap{\rule{\dimexpr\labelwidth+\labelsep}{\phaserulewidth}}\rule{\linewidth}{\phaserulewidth}}
\makeatother
\setphaserulewidth{.7pt}

\maketitle

\newcommand{\figleft}{{\em (Left)}}
\newcommand{\figcenter}{{\em (Center)}}
\newcommand{\figright}{{\em (Right)}}
\newcommand{\figtop}{{\em (Top)}}
\newcommand{\figbottom}{{\em (Bottom)}}
\newcommand{\captiona}{{\em (a)}}
\newcommand{\captionb}{{\em (b)}}
\newcommand{\captionc}{{\em (c)}}
\newcommand{\captiond}{{\em (d)}}

\newcommand{\newterm}[1]{{\bf #1}}

\def\figref#1{figure~\ref{#1}}
\def\Figref#1{Figure~\ref{#1}}
\def\twofigref#1#2{figures \ref{#1} and \ref{#2}}
\def\quadfigref#1#2#3#4{figures \ref{#1}, \ref{#2}, \ref{#3} and \ref{#4}}
\def\secref#1{section~\ref{#1}}
\def\Secref#1{Section~\ref{#1}}
\def\twosecrefs#1#2{sections \ref{#1} and \ref{#2}}
\def\secrefs#1#2#3{sections \ref{#1}, \ref{#2} and \ref{#3}}
\def\eqref#1{equation~\ref{#1}}
\def\Eqref#1{Equation~\ref{#1}}
\def\plaineqref#1{\ref{#1}}
\def\chapref#1{chapter~\ref{#1}}
\def\Chapref#1{Chapter~\ref{#1}}
\def\rangechapref#1#2{chapters\ref{#1}--\ref{#2}}
\def\algref#1{algorithm~\ref{#1}}
\def\Algref#1{Algorithm~\ref{#1}}
\def\twoalgref#1#2{algorithms \ref{#1} and \ref{#2}}
\def\Twoalgref#1#2{Algorithms \ref{#1} and \ref{#2}}
\def\partref#1{part~\ref{#1}}
\def\Partref#1{Part~\ref{#1}}
\def\twopartref#1#2{parts \ref{#1} and \ref{#2}}

\def\ceil#1{\lceil #1 \rceil}
\def\floor#1{\lfloor #1 \rfloor}
\def\1{\bm{1}}
\newcommand{\train}{\mathcal{D}}
\newcommand{\valid}{\mathcal{D_{\mathrm{valid}}}}
\newcommand{\test}{\mathcal{D_{\mathrm{test}}}}

\def\eps{{\epsilon}}

\def\reta{{\textnormal{$\eta$}}}
\def\ra{{\textnormal{a}}}
\def\rb{{\textnormal{b}}}
\def\rc{{\textnormal{c}}}
\def\rd{{\textnormal{d}}}
\def\re{{\textnormal{e}}}
\def\rf{{\textnormal{f}}}
\def\rg{{\textnormal{g}}}
\def\rh{{\textnormal{h}}}
\def\ri{{\textnormal{i}}}
\def\rj{{\textnormal{j}}}
\def\rk{{\textnormal{k}}}
\def\rl{{\textnormal{l}}}
\def\rn{{\textnormal{n}}}
\def\ro{{\textnormal{o}}}
\def\rp{{\textnormal{p}}}
\def\rq{{\textnormal{q}}}
\def\rr{{\textnormal{r}}}
\def\rs{{\textnormal{s}}}
\def\rt{{\textnormal{t}}}
\def\ru{{\textnormal{u}}}
\def\rv{{\textnormal{v}}}
\def\rw{{\textnormal{w}}}
\def\rx{{\textnormal{x}}}
\def\ry{{\textnormal{y}}}
\def\rz{{\textnormal{z}}}

\def\rvepsilon{{\mathbf{\epsilon}}}
\def\rvtheta{{\mathbf{\theta}}}
\def\rva{{\mathbf{a}}}
\def\rvb{{\mathbf{b}}}
\def\rvc{{\mathbf{c}}}
\def\rvd{{\mathbf{d}}}
\def\rve{{\mathbf{e}}}
\def\rvf{{\mathbf{f}}}
\def\rvg{{\mathbf{g}}}
\def\rvh{{\mathbf{h}}}
\def\rvu{{\mathbf{i}}}
\def\rvj{{\mathbf{j}}}
\def\rvk{{\mathbf{k}}}
\def\rvl{{\mathbf{l}}}
\def\rvm{{\mathbf{m}}}
\def\rvn{{\mathbf{n}}}
\def\rvo{{\mathbf{o}}}
\def\rvp{{\mathbf{p}}}
\def\rvq{{\mathbf{q}}}
\def\rvr{{\mathbf{r}}}
\def\rvs{{\mathbf{s}}}
\def\rvt{{\mathbf{t}}}
\def\rvu{{\mathbf{u}}}
\def\rvv{{\mathbf{v}}}
\def\rvw{{\mathbf{w}}}
\def\rvx{{\mathbf{x}}}
\def\rvy{{\mathbf{y}}}
\def\rvz{{\mathbf{z}}}

\def\erva{{\textnormal{a}}}
\def\ervb{{\textnormal{b}}}
\def\ervc{{\textnormal{c}}}
\def\ervd{{\textnormal{d}}}
\def\erve{{\textnormal{e}}}
\def\ervf{{\textnormal{f}}}
\def\ervg{{\textnormal{g}}}
\def\ervh{{\textnormal{h}}}
\def\ervi{{\textnormal{i}}}
\def\ervj{{\textnormal{j}}}
\def\ervk{{\textnormal{k}}}
\def\ervl{{\textnormal{l}}}
\def\ervm{{\textnormal{m}}}
\def\ervn{{\textnormal{n}}}
\def\ervo{{\textnormal{o}}}
\def\ervp{{\textnormal{p}}}
\def\ervq{{\textnormal{q}}}
\def\ervr{{\textnormal{r}}}
\def\ervs{{\textnormal{s}}}
\def\ervt{{\textnormal{t}}}
\def\ervu{{\textnormal{u}}}
\def\ervv{{\textnormal{v}}}
\def\ervw{{\textnormal{w}}}
\def\ervx{{\textnormal{x}}}
\def\ervy{{\textnormal{y}}}
\def\ervz{{\textnormal{z}}}

\def\rmA{{\mathbf{A}}}
\def\rmB{{\mathbf{B}}}
\def\rmC{{\mathbf{C}}}
\def\rmD{{\mathbf{D}}}
\def\rmE{{\mathbf{E}}}
\def\rmF{{\mathbf{F}}}
\def\rmG{{\mathbf{G}}}
\def\rmH{{\mathbf{H}}}
\def\rmI{{\mathbf{I}}}
\def\rmJ{{\mathbf{J}}}
\def\rmK{{\mathbf{K}}}
\def\rmL{{\mathbf{L}}}
\def\rmM{{\mathbf{M}}}
\def\rmN{{\mathbf{N}}}
\def\rmO{{\mathbf{O}}}
\def\rmP{{\mathbf{P}}}
\def\rmQ{{\mathbf{Q}}}
\def\rmR{{\mathbf{R}}}
\def\rmS{{\mathbf{S}}}
\def\rmT{{\mathbf{T}}}
\def\rmU{{\mathbf{U}}}
\def\rmV{{\mathbf{V}}}
\def\rmW{{\mathbf{W}}}
\def\rmX{{\mathbf{X}}}
\def\rmY{{\mathbf{Y}}}
\def\rmZ{{\mathbf{Z}}}

\def\ermA{{\textnormal{A}}}
\def\ermB{{\textnormal{B}}}
\def\ermC{{\textnormal{C}}}
\def\ermD{{\textnormal{D}}}
\def\ermE{{\textnormal{E}}}
\def\ermF{{\textnormal{F}}}
\def\ermG{{\textnormal{G}}}
\def\ermH{{\textnormal{H}}}
\def\ermI{{\textnormal{I}}}
\def\ermJ{{\textnormal{J}}}
\def\ermK{{\textnormal{K}}}
\def\ermL{{\textnormal{L}}}
\def\ermM{{\textnormal{M}}}
\def\ermN{{\textnormal{N}}}
\def\ermO{{\textnormal{O}}}
\def\ermP{{\textnormal{P}}}
\def\ermQ{{\textnormal{Q}}}
\def\ermR{{\textnormal{R}}}
\def\ermS{{\textnormal{S}}}
\def\ermT{{\textnormal{T}}}
\def\ermU{{\textnormal{U}}}
\def\ermV{{\textnormal{V}}}
\def\ermW{{\textnormal{W}}}
\def\ermX{{\textnormal{X}}}
\def\ermY{{\textnormal{Y}}}
\def\ermZ{{\textnormal{Z}}}

\def\vzero{{\bm{0}}}
\def\vone{{\bm{1}}}
\def\vmu{{\bm{\mu}}}
\def\vtheta{{\bm{\theta}}}
\def\va{{\bm{a}}}
\def\vb{{\bm{b}}}
\def\vc{{\bm{c}}}
\def\vd{{\bm{d}}}
\def\ve{{\bm{e}}}
\def\vf{{\bm{f}}}
\def\vg{{\bm{g}}}
\def\vh{{\bm{h}}}
\def\vi{{\bm{i}}}
\def\vj{{\bm{j}}}
\def\vk{{\bm{k}}}
\def\vl{{\bm{l}}}
\def\vm{{\bm{m}}}
\def\vn{{\bm{n}}}
\def\vo{{\bm{o}}}
\def\vp{{\bm{p}}}
\def\vq{{\bm{q}}}
\def\vr{{\bm{r}}}
\def\vs{{\bm{s}}}
\def\vt{{\bm{t}}}
\def\vu{{\bm{u}}}
\def\vv{{\bm{v}}}
\def\vw{{\bm{w}}}
\def\vx{{\bm{x}}}
\def\vy{{\bm{y}}}
\def\vz{{\bm{z}}}

\def\evalpha{{\alpha}}
\def\evbeta{{\beta}}
\def\evepsilon{{\epsilon}}
\def\evlambda{{\lambda}}
\def\evomega{{\omega}}
\def\evmu{{\mu}}
\def\evpsi{{\psi}}
\def\evsigma{{\sigma}}
\def\evtheta{{\theta}}
\def\eva{{a}}
\def\evb{{b}}
\def\evc{{c}}
\def\evd{{d}}
\def\eve{{e}}
\def\evf{{f}}
\def\evg{{g}}
\def\evh{{h}}
\def\evi{{i}}
\def\evj{{j}}
\def\evk{{k}}
\def\evl{{l}}
\def\evm{{m}}
\def\evn{{n}}
\def\evo{{o}}
\def\evp{{p}}
\def\evq{{q}}
\def\evr{{r}}
\def\evs{{s}}
\def\evt{{t}}
\def\evu{{u}}
\def\evv{{v}}
\def\evw{{w}}
\def\evx{{x}}
\def\evy{{y}}
\def\evz{{z}}

\def\mA{{\bm{A}}}
\def\mB{{\bm{B}}}
\def\mC{{\bm{C}}}
\def\mD{{\bm{D}}}
\def\mE{{\bm{E}}}
\def\mF{{\bm{F}}}
\def\mG{{\bm{G}}}
\def\mH{{\bm{H}}}
\def\mI{{\bm{I}}}
\def\mJ{{\bm{J}}}
\def\mK{{\bm{K}}}
\def\mL{{\bm{L}}}
\def\mM{{\bm{M}}}
\def\mN{{\bm{N}}}
\def\mO{{\bm{O}}}
\def\mP{{\bm{P}}}
\def\mQ{{\bm{Q}}}
\def\mR{{\bm{R}}}
\def\mS{{\bm{S}}}
\def\mT{{\bm{T}}}
\def\mU{{\bm{U}}}
\def\mV{{\bm{V}}}
\def\mW{{\bm{W}}}
\def\mX{{\bm{X}}}
\def\mY{{\bm{Y}}}
\def\mZ{{\bm{Z}}}
\def\mBeta{{\bm{\beta}}}
\def\mPhi{{\bm{\Phi}}}
\def\mLambda{{\bm{\Lambda}}}
\def\mSigma{{\bm{\Sigma}}}

\newcommand{\tens}[1]{\bm{\mathsfit{#1}}}
\def\tA{{\tens{A}}}
\def\tB{{\tens{B}}}
\def\tC{{\tens{C}}}
\def\tD{{\tens{D}}}
\def\tE{{\tens{E}}}
\def\tF{{\tens{F}}}
\def\tG{{\tens{G}}}
\def\tH{{\tens{H}}}
\def\tI{{\tens{I}}}
\def\tJ{{\tens{J}}}
\def\tK{{\tens{K}}}
\def\tL{{\tens{L}}}
\def\tM{{\tens{M}}}
\def\tN{{\tens{N}}}
\def\tO{{\tens{O}}}
\def\tP{{\tens{P}}}
\def\tQ{{\tens{Q}}}
\def\tR{{\tens{R}}}
\def\tS{{\tens{S}}}
\def\tT{{\tens{T}}}
\def\tU{{\tens{U}}}
\def\tV{{\tens{V}}}
\def\tW{{\tens{W}}}
\def\tX{{\tens{X}}}
\def\tY{{\tens{Y}}}
\def\tZ{{\tens{Z}}}

\def\gA{{\mathcal{A}}}
\def\gB{{\mathcal{B}}}
\def\gC{{\mathcal{C}}}
\def\gD{{\mathcal{D}}}
\def\gE{{\mathcal{E}}}
\def\gF{{\mathcal{F}}}
\def\gG{{\mathcal{G}}}
\def\gH{{\mathcal{H}}}
\def\gI{{\mathcal{I}}}
\def\gJ{{\mathcal{J}}}
\def\gK{{\mathcal{K}}}
\def\gL{{\mathcal{L}}}
\def\gM{{\mathcal{M}}}
\def\gN{{\mathcal{N}}}
\def\gO{{\mathcal{O}}}
\def\gP{{\mathcal{P}}}
\def\gQ{{\mathcal{Q}}}
\def\gR{{\mathcal{R}}}
\def\gS{{\mathcal{S}}}
\def\gT{{\mathcal{T}}}
\def\gU{{\mathcal{U}}}
\def\gV{{\mathcal{V}}}
\def\gW{{\mathcal{W}}}
\def\gX{{\mathcal{X}}}
\def\gY{{\mathcal{Y}}}
\def\gZ{{\mathcal{Z}}}

\def\sA{{\mathbb{A}}}
\def\sB{{\mathbb{B}}}
\def\sC{{\mathbb{C}}}
\def\sD{{\mathbb{D}}}
\def\sF{{\mathbb{F}}}
\def\sG{{\mathbb{G}}}
\def\sH{{\mathbb{H}}}
\def\sI{{\mathbb{I}}}
\def\sJ{{\mathbb{J}}}
\def\sK{{\mathbb{K}}}
\def\sL{{\mathbb{L}}}
\def\sM{{\mathbb{M}}}
\def\sN{{\mathbb{N}}}
\def\sO{{\mathbb{O}}}
\def\sP{{\mathbb{P}}}
\def\sQ{{\mathbb{Q}}}
\def\sR{{\mathbb{R}}}
\def\sS{{\mathbb{S}}}
\def\sT{{\mathbb{T}}}
\def\sU{{\mathbb{U}}}
\def\sV{{\mathbb{V}}}
\def\sW{{\mathbb{W}}}
\def\sX{{\mathbb{X}}}
\def\sY{{\mathbb{Y}}}
\def\sZ{{\mathbb{Z}}}

\def\emLambda{{\Lambda}}
\def\emA{{A}}
\def\emB{{B}}
\def\emC{{C}}
\def\emD{{D}}
\def\emE{{E}}
\def\emF{{F}}
\def\emG{{G}}
\def\emH{{H}}
\def\emI{{I}}
\def\emJ{{J}}
\def\emK{{K}}
\def\emL{{L}}
\def\emM{{M}}
\def\emN{{N}}
\def\emO{{O}}
\def\emP{{P}}
\def\emQ{{Q}}
\def\emR{{R}}
\def\emS{{S}}
\def\emT{{T}}
\def\emU{{U}}
\def\emV{{V}}
\def\emW{{W}}
\def\emX{{X}}
\def\emY{{Y}}
\def\emZ{{Z}}
\def\emSigma{{\Sigma}}

\newcommand{\etens}[1]{\mathsfit{#1}}
\def\etLambda{{\etens{\Lambda}}}
\def\etA{{\etens{A}}}
\def\etB{{\etens{B}}}
\def\etC{{\etens{C}}}
\def\etD{{\etens{D}}}
\def\etE{{\etens{E}}}
\def\etF{{\etens{F}}}
\def\etG{{\etens{G}}}
\def\etH{{\etens{H}}}
\def\etI{{\etens{I}}}
\def\etJ{{\etens{J}}}
\def\etK{{\etens{K}}}
\def\etL{{\etens{L}}}
\def\etM{{\etens{M}}}
\def\etN{{\etens{N}}}
\def\etO{{\etens{O}}}
\def\etP{{\etens{P}}}
\def\etQ{{\etens{Q}}}
\def\etR{{\etens{R}}}
\def\etS{{\etens{S}}}
\def\etT{{\etens{T}}}
\def\etU{{\etens{U}}}
\def\etV{{\etens{V}}}
\def\etW{{\etens{W}}}
\def\etX{{\etens{X}}}
\def\etY{{\etens{Y}}}
\def\etZ{{\etens{Z}}}

\newcommand{\pdata}{p_{\rm{data}}}
\newcommand{\ptrain}{\hat{p}_{\rm{data}}}
\newcommand{\Ptrain}{\hat{P}_{\rm{data}}}
\newcommand{\pmodel}{p_{\rm{model}}}
\newcommand{\Pmodel}{P_{\rm{model}}}
\newcommand{\ptildemodel}{\tilde{p}_{\rm{model}}}
\newcommand{\pencode}{p_{\rm{encoder}}}
\newcommand{\pdecode}{p_{\rm{decoder}}}
\newcommand{\precons}{p_{\rm{reconstruct}}}

\newcommand{\laplace}{\mathrm{Laplace}} 

\newcommand{\E}{\mathbb{E}}
\newcommand{\Ls}{\mathcal{L}}
\newcommand{\R}{\mathbb{R}}
\newcommand{\emp}{\tilde{p}}
\newcommand{\lr}{\alpha}
\newcommand{\reg}{\lambda}
\newcommand{\rect}{\mathrm{rectifier}}
\newcommand{\softmax}{\mathrm{softmax}}
\newcommand{\sigmoid}{\sigma}
\newcommand{\softplus}{\zeta}
\newcommand{\KL}{D_{\mathrm{KL}}}
\newcommand{\Var}{\mathrm{Var}}
\newcommand{\standarderror}{\mathrm{SE}}
\newcommand{\Cov}{\mathrm{Cov}}
\newcommand{\normlzero}{L^0}
\newcommand{\normlone}{L^1}
\newcommand{\normltwo}{L^2}
\newcommand{\normlp}{L^p}
\newcommand{\normmax}{L^\infty}

\newcommand{\parents}{Pa} 

\let\ab\allowbreak

\newcommand{\bA}{\mathbf{A}}
\newcommand{\bI}{\mathbf{I}}
\newcommand{\bJ}{\mathbf{J}}
\newcommand{\bH}{\mathbf{H}}
\newcommand{\bL}{\mathbf{L}}
\newcommand{\bM}{\mathbf{M}}
\newcommand{\bQ}{\mathbf{Q}}
\newcommand{\bR}{\mathbf{R}}
\newcommand{\bzero}{\mathbf{0}}
\newcommand{\bone}{\mathbf{1}}
\newcommand{\bb}{\mathbf{b}}
\newcommand{\bu}{\mathbf{u}}
\newcommand{\bv}{\mathbf{v}}
\newcommand{\bw}{\mathbf{w}}
\newcommand{\bx}{\mathbf{x}}
\newcommand{\by}{\mathbf{y}}
\newcommand{\bz}{\mathbf{z}}
\newcommand{\bh}{\mathbf{h}}
\newcommand{\be}{\mathbf{e}}
\newcommand{\bq}{\mathbf{q}}
\newcommand{\bk}{\mathbf{k}}
\newcommand{\bxh}{\hat{\mathbf{x}}}
\newcommand{\btheta}{{\mathbf{\theta}}}
\newcommand{\bphi}{{\mathbf{\phi}}}
\newcommand{\bepsilon}{{\mathbf{\epsilon}}}
\newcommand{\bmu}{{\mathbf{\mu}}}
\newcommand{\bnu}{{\mathbf{\nu}}}
\newcommand{\bSigma}{{\mathbf{\Sigma}}}
\newcommand{\defeq}{\coloneqq}

\begin{abstract}
Video and audio content creation serves as the core technique for the movie industry and professional users. Recently, existing diffusion-based methods tackle video and audio generation separately, which hinders the technique transfer from academia to industry. In this work, we aim at filling the gap, with a carefully designed optimization-based framework for cross-visual-audio and joint-visual-audio generation. We observe the powerful generation ability of off-the-shelf video or audio generation models. Thus, instead of training the giant models from scratch, we propose to bridge the existing strong models with a shared latent representation space. 
Specifically, we propose a multimodality latent aligner with the pre-trained ImageBind model. Our latent aligner shares a similar core as the classifier guidance that guides the diffusion denoising process during inference time. Through carefully designed optimization strategy and loss functions, we show the superior performance of our method on joint video-audio generation, visual-steered audio generation, and audio-steered visual generation tasks. The project website can be found at~\href{https://yzxing87.github.io/Seeing-and-Hearing/}{https://yzxing87.github.io/Seeing-and-Hearing/}. 
\end{abstract}    
\begin{figure*}[t]
    \centering
    \includegraphics[width=\linewidth]{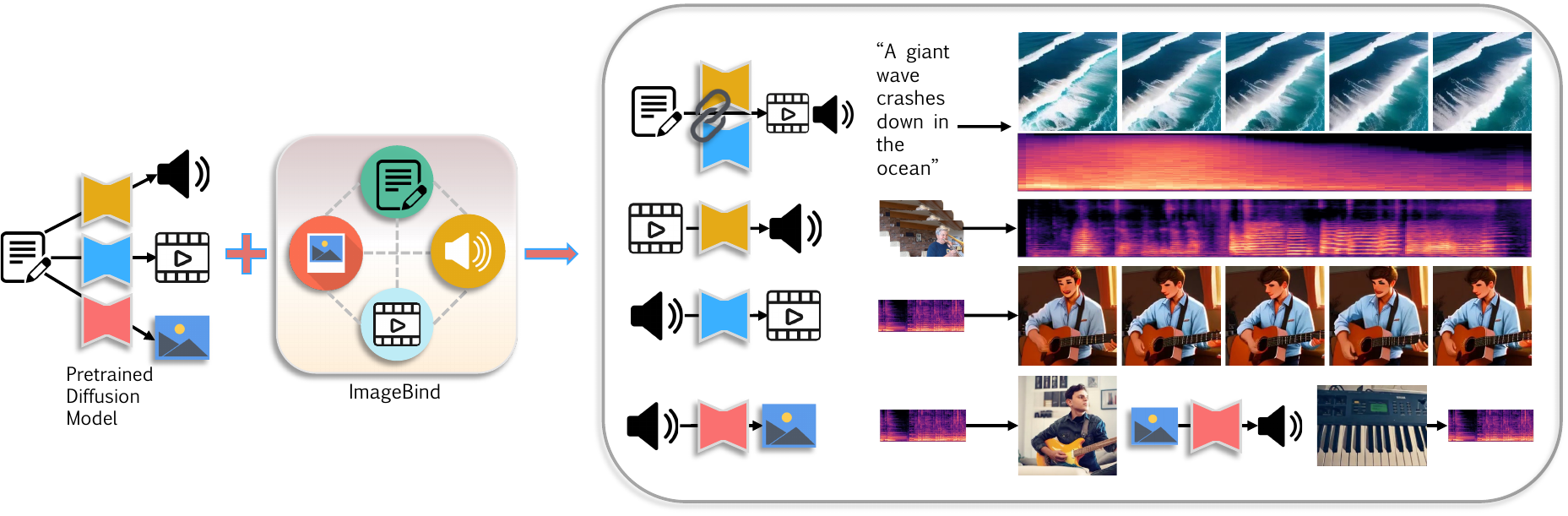}
    \caption{
    \textbf{Overview.}
    Our approach is versatile and can tackle four tasks: joint video-audio generation (Joint-VA), video-to-audio (V2A), audio-to-video (A2V), and image-to-audio (I2A).
    By leveraging a multimodal binder, e.g., pretrained ImageBind, 
    we establish a connection between isolated generative models that are designed for generating a single modality. 
    This enables us to achieve both bidirectional conditional and joint video/audio generation.
    }
    \label{fig:attention}
    \vspace{-3mm}
\end{figure*}

\section{Introduction}
\label{sec:intro}

Recently, AI-generated content has made significant advances in creating diverse and high-realistic images~\cite{dai2023emu, ramesh2022dalle2, ho2022imagen, videoldm, nichol2021glide}, videos~\cite{ho2022imagen,videoldm,make-a-video,he2022latent,chen2023videocrafter,pyoco,he2023animate}, or sound~\cite{kreuk2022audiogen,yang2023diffsound, huang2023make, liu2023audioldm,liu2023audioldm2}, based on the input descriptions from users. 
However, existing works primarily concentrate on generating content within a single modality, disregarding the multimodal nature of the real world.
Consequently, the generated videos lack accompanying audio, and the generated audio lacks synchronized visual effects.
This research gap restricts users from creating content with greater impact, such as producing films that necessitate the simultaneous creation of both visual and audio modalities.
In this work, we study the visual-audio generation task for crafting both video and audio content.

One potential solution to this problem is to generate visual and audio content in two stages.
For example, users can first generate the video based on the input text prompt utilizing existing text-to-video (T2V) models~\cite{chen2023videocrafter,guo2023animatediff}.
Then, a video-to-audio (V2A) model can be employed to generate aligned audio. 
Alternatively, a combination of text-to-audio (T2A) and audio-to-video (A2V) models can be used to generate paired visual-audio content.
However, existing V2A and A2V generation methods~\cite{iashin2021specvqgan, yariv2023tempotokens} either have limited capability to specific downstream domains or exhibit poor generation performance. 
Moreover, the task of joint video-audio generation (Joint-VA) has received limited attention, and existing work~\cite{ruan2023mm} shows limited generation performance even within a small domain and also lacks semantic control.

In this work, we propose a new generation paradigm for open-domain visual-audio generation.
We observe that:
(1) There are well-trained single-modality text-conditioned generation models that demonstrate excellent performance. Leveraging these pre-trained models can avoid expensive training for synthesizing each modality.
(2) We have noticed that the pre-trained model ImageBind~\cite{girdhar2023imagebind} possesses remarkable capability in establishing effective connections between different data modalities within a shared semantic space. 
Our objective is to explore how we can leverage ImageBind as a bridge to connect and integrate various modalities effectively.

Leveraging these observations, we propose to utilize ImageBind as an aligner in the diffusion latent space of different modalities.
During the generation of one modality, we input the noisy latent and the guided condition of another modality to our aligner to produce a guidance signal that influences the generation process.
By gradually injecting the guidance into the denoising process, we bridge the generated content closer to the input condition in the ImageBind embedding space.
For Joint-VA generation, we make the guidance bidirectional to impact the generation processes of both modalities.

With our design, we successfully bridge the pre-trained single-modality generation models into an organic system and achieve a versatile and flexible visual-audio generation.
In addition, our approach does not require training on large-scale datasets, making our approach very resource-friendly.
Besides the generality and low cost of our approach, we validate our performance on four tasks and show the superiority over baseline approaches. 

In summary, our key contributions are as follows:
\begin{itemize}
    \item We propose a novel paradigm that \textit{bridges} pre-trained diffusion models of single modality together to achieve audio-visual generation.
    \item We introduce \textit{diffusion latent aligner} to gradually align diffusion latent of visual and audio modalities in a multimodal embedding space.
    \item We conduct extensive experiments on four tasks including V2A, I2A, A2V, and Joint-VA, demonstrating the superiority and generality of our approach.
    \item  To the best of our knowledge, we present the first work for text-guided joint video-audio generation.
\end{itemize}

\section{Related Work}
\label{sec:related}

\subsection{Conditional Audio Generation}
Audio generation is an emerging field that focuses on modeling the creation of diverse audio content. This includes tasks such as generating audio conditioned on various inputs like text~\cite{kreuk2022audiogen,yang2023diffsound, huang2023make,huang2023make2,ghosal2023TANGO,dong2023clipsonic}, images~\cite{sheffer2023im2wav}, and videos~\cite{iashin2021specvqgan,luo2023diff-foley, du2023conditional, su2023physics}.

In the field of text-to-audio research, AudioGen~\cite{kreuk2022audiogen} proposes an auto-regressive generative model that operates on discrete audio representations, DiffSound~\cite{yang2023diffsound} utilizes non-autoregressive token-decoder to address the limitations of unidirectional generation in auto-regressive models. While some other works like Make-An-Audio~\cite{huang2023make}, AudioLDM~\cite{liu2023audioldm}, employ latent diffusion methods for audio generation. Some recent studies, such as Make-an-Audio2~\cite{huang2023make2}, AudioLDM2~\cite{liu2023audioldm2}, TANGO~\cite{ghosal2023TANGO}, have leveraged Large Language Models (LLMs) to enhance the performance of audio generation models.

Research focusing on audio generation that is conditioned on images and videos, exemplified by works like Im2Wav~\cite{sheffer2023im2wav} and SpecVQGAN~\cite{iashin2021specvqgan}, has also captured significant interest within the scholarly community. Utilizing the semantics of a pre-trained CLIP model for visual representation (Contrastive Language–Image Pre-training)~\cite{radford2021clip}, Im2Wav~\cite{sheffer2023im2wav} first crafts a foundational audio representation via a language model, then employs an additional language model to upsample these audio tokens into high-fidelity sound samples. SpecVQGAN~\cite{iashin2021specvqgan} utilizes a transformer to generate new spectrograms from a pre-trained codebook based on input video features. It then reconstructs the waveform from these spectrograms using a pre-trained vocoder.

\subsection{Conditional Visual Generation}
The task of text-to-image generation has seen significant development and achievements in recent years~\cite{tao2023galipgan, avrahami2023spatext, RombachBLEO22-stable-diff}. This progress has sparked interest in a new research domain focusing on audio-to-image generation.
In 2019, ~\cite{wan2019towards} proposed a method to generate images from audio recordings, employing Generative Adversarial Networks (GANs). ~\cite{zelaszczyk2022audio} focused narrowly on generating images of MNIST digits using audio inputs and did not extend to image generation from general audio sounds. In contrast, the approach by ~\cite{wan2019towards}, was capable of generating images from a broader range of audio signals. Wav2CLIP~\cite{wu2022wav2clip} adopts a CLIP-inspired approach to learn joint representations for audio-image pairs, which can subsequently facilitate image generation using VQ-GAN~\cite{esser2021taming}.
Text-to-video has also achieved remarkable progress recently~\cite{ho2022video, he2022latent, zhou2022magicvideo, ho2022imagen, huang2023make,videoldm,pyoco,zhang2023show,an2023latent-shift,chen2023videocrafter} empowered by video diffusion models~\cite{ho2022video}.
The mainstream idea is to incorporate temporal modeling modules in the U-Net architecture to learn the temporal dynamics~\cite{ho2022video, make-a-video, he2022latent, zhou2022magicvideo,an2023latent-shift} in the video pixel space~\cite{ho2022video, ho2022imagen} or in the latent space~\cite{videoldm,he2022latent}. 
In this work, we leverage the open-source latent-based text-to-video model as our base model for the video generation counterpart.
There're also some Audio-to-video works that have been done, such as Sound2sight~\cite{chatterjee2020sound2sight}, TATS~\cite{ge2022long}, and Tempotokens~\cite{yariv2023diverse}. While~\cite{chatterjee2020sound2sight} focuses on extending videos in a way that aligns with the audio, Tempotokens~\cite{yariv2023diverse} takes a different approach by exclusively generating videos from the audio input. TATS~\cite{ge2022long} introduced a technique for creating videos synchronized with audio, but despite its remarkable aspects, the variety in the videos it produces is significantly constrained.

\subsection{Multimodal Joint Generation}
Some research has already begun exploring the area of Multimodal Joint Generation~\cite{ruan2023mm, zhu2023moviefactory}. MM-Diffusion~\cite{ruan2023mm} introduces the first framework for simultaneous audio-video generation, designed to synergistically enhance both visual and auditory experiences cohesively and engagingly. However, it's unconditional and can only generate results in the training set domain, which would limit generation diversity. MovieFactory~\cite{zhu2023moviefactory} employs ChatGPT to elaborately expand user-input text into detailed sequential scripts for generating movies, which are then vividly actualized both visually and acoustically through vision generation and audio retrieval techniques. However, a notable constraint of MovieFactory lies in its reliance on audio retrieval, limiting its capacity to generate audio that is more intricately tailored to the specific scenes.

\section{Method}

\begin{figure*}[t]
    \centering
    \includegraphics[width=0.9\linewidth]{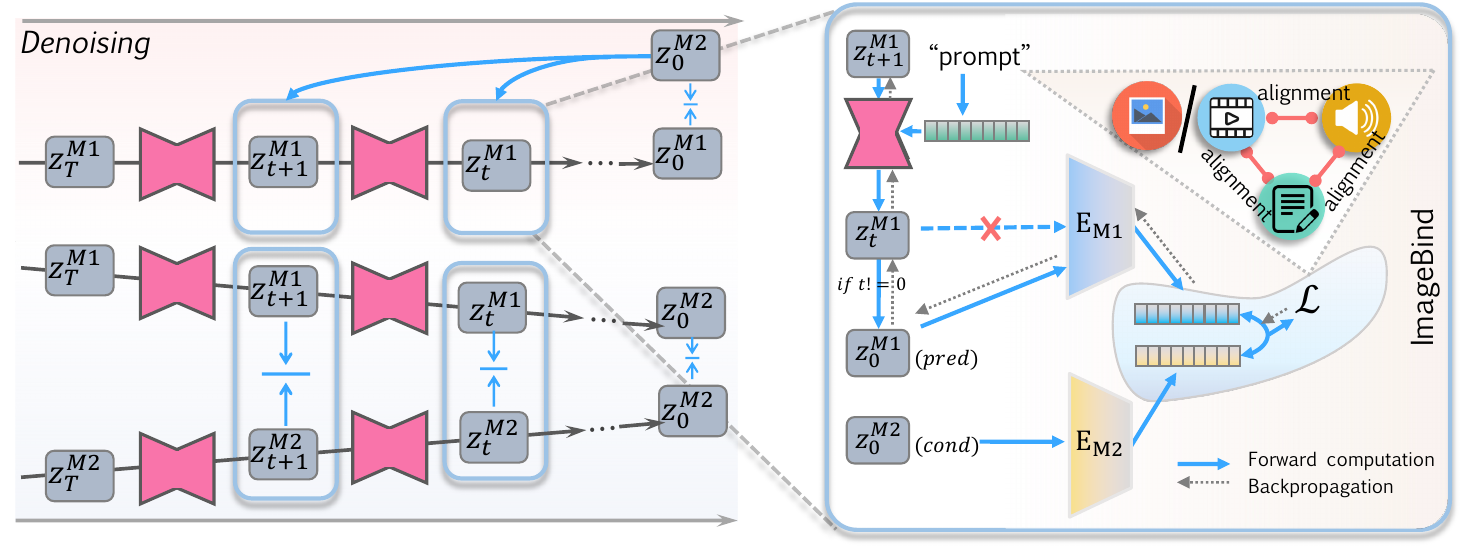}
    \vspace{-3mm}
    \caption{\textbf{The proposed diffusion latent aligner.}
        During the denoising process of generating one specific modality (visual/audio), we adopt the condition information (audio/video) to guide the denoising process.
        By leveraging the pretrained ImageBind model, we calculate the distance of the generative latent $\bz_{t}^{M_1}$ with the condition $\bz_0^{M_2}$ in the shared embedding space of ImageBind.
        Then we backpropagate the distance value to obtain the gradient of $\bz_{t}^{M_1}$ with respect to the distance.}
    \label{fig:method}
    \vspace{-3mm}
\end{figure*}

\subsection{Preliminaries}
\subsubsection{Latent diffusion models}
We adopt latent-based diffusion models (LDM) for our generation model. 
The diffusion process follows the standard formulation in DDPM \cite{ho2020denoising} that consists of a forward diffusion and a backward denoising process.
Given a data sample $\mathbf{x} \sim p(\mathbf{x})$, an autoencoder consisting an encoder $\mathcal{E}$ and a decoder $\mathcal{D}$ first project the $\mathbf{x}$ into latent $\mathbf{z}$ via $\mathbf{z}= \mathcal{E}(\mathbf{x})$.
Then, the diffusion and denoising process are conducted in the latent space. 
Once the denoising is completed at timestep 0, the sample will be decoded via $\bx = \mathcal{D}(\tilde{\bz_0})$.
The forward diffusion is a fixed Markov process of $T$ timesteps that yields latent variables $\mathbf{z}_t$ based on the latent variable at previous timestep $\mathbf{z}_{t-1}$ via
\begin{equation}
    q(\bz_t|\bz_{t-1}) = \mathcal{N}(\bz_t;\sqrt{1-\beta_t}\bz_{t-1},\beta_t \bI), \\ 
\end{equation}
where $\beta_t$ is a predefined variance at each step $t$.
Finally, the clean data $\bz_0$ becomes $\bz_T$, which is indistinguishable from a Gaussian noise.
The $\bz_t$ can be directly derived from $\bz_0$ in a closed form:
\begin{equation}
    q(\bz_t|\bz_0) = \mathcal{N}(\bz_t; \sqrt{\bar\alpha_t}\bz_0, (1-\bar\alpha_t)\bI), 
\end{equation}
where $\bar\alpha_t = \prod_{i=1}^t \alpha_i$, and $\alpha_t = 1-\beta_t$.
Leveraging the reparameterization trick, the $\bz_t$ can be computed via
\begin{equation}
    \bz_t = \sqrt{\bar\alpha_t}\bz_0 + (1-\bar\alpha_t)\mathbf\epsilon,
\end{equation}
where $\mathbf\epsilon$ is a random Gaussian noise.
The backward denoising process leverages a trained denoiser $\theta$ to obtain less noisy data $\mathbf{z}_{t-1}$ from the noisy input $\mathbf{z}_t$ at each timestep:
\begin{equation}
p_\theta\left(\mathbf{z}_{t-1} \mid \mathbf{z}_t\right)=\mathcal{N}\left(\mathbf{z}_{t-1} ; \mathbf{\mu}_\theta\left(\mathbf{z}_t, t, p\right), \mathbf{\Sigma}_\theta\left(\mathbf{z}_t, t, p\right)\right).
\end{equation}

Here $\mathbf{\mu}_\theta$ and $\mathbf{\Sigma}_\theta$ are determined through a denoiser network $\mathbf{\epsilon}_{\theta}\left(\mathbf{z}_t, t, p\right)$, where $p$ represents input prompt.
The training objective of $\theta$ is a noise estimation loss, formulated as
\begin{equation}
\min _{\theta} \mathbb{E}_{t, \mathbf{z}_t, \mathbf{\epsilon}}\left\|\mathbf{\epsilon}-\mathbf{\epsilon}_{\theta}\left(\mathbf{z}_t, t, p\right)\right\|_2^2.
\end{equation}

\subsubsection{Classifier guidance}
Classifier guidance~\cite{dhariwal2021guidediffusion} is a conditional generation mechanism that leverages the unconditional diffusion model to generate samples with the desired category.
Given an unconditional diffusion model $p_\theta(\bz_t|\bz_{t+1})$, in order to condition it on a class label $y$, it can be approximated via
\begin{align}
    p_{\theta, \phi}(\bz_t|\bz_{t+1}, y) = \mathcal{Z}p_\theta(\bz_t|\bz_{t+1})p_\phi(y|\bz_t, t),
\end{align}
where $\mathcal{Z}$ is a constant coefficient for normalization, $\phi$ is a trained time-aware noisy classifier for the approximation of label distribution of each sample of $\bz_t$.
The guidance from the classifier $\phi$ is the gradient of $\bz_t$ with respect to y and is applied to the original $\bz_t$ predicted from $\epsilon_\theta$:
\begin{align}
\hat{\epsilon}(\bz_t) = \epsilon_\theta(\bz_t) - \sqrt{1-\hat{\alpha}_t} \triangledown_{\bz_t} \log p_{\phi}(y|\bz_t).
\end{align}

\subsubsection{Linking multiple modalities}
We aim to force the generated samples in different modalities to become closer in a joint semantic space.
To achieve this goal, we choose ImageBind~\cite{girdhar2023imagebind} as the aligner since it learns an effective joint embedding space for multiple modalities.
ImageBind learns a joint semantic embedding space that binds multiple different modalities including image, text, video, audio, depth, and thermal.
Given a pair of data with different modalities ($M_1, M_2$), e.g., (video, audio), the encoder of the corresponding modality $\mathbf{E}_i$ takes the data as input and predicts its embedding $\be_i$.
The ImageBind is trained with a contrastive learning objective formulated as follows:
\begin{equation}
    \mathcal{L}_{M_1,M_2} = - \log \frac{\exp(\bq_{i}^{\intercal} \bk_i/\tau)}{\exp(\bq_{i}^{\intercal} \bk_i/\tau) + \sum_{j \neq i}\exp(\bq_{i}^{\intercal} \bk_j/\tau)},
    \label{eq:contrastive_loss}
\end{equation}
where $\tau$ is a temperature factor to control the smoothness of the Softmax distribution, and $j$ represents the negative sample, which is the data from another pair.
By projecting samples of different modalities into embeddings in a shared space, minimizing the distance of the embeddings from the same data pair, and maximizing the distance of the embeddings from different data pairs, the ImageBind model achieves semantic alignment capability and thus can be served as a desired tool for multimodal alignment.

\subsection{Diffusion Latent Aligner}
\subsubsection{Problem formulation}
\label{sec:formulation}
Consider two modalities ${M_1, M_2}$, where $M_2$ is the conditional modality and $M_1$ is the generative modality. 
Given a latent diffusion model (LDM) $\theta$ that produces data of $M_1$, our objective is to leverage the information from the condition $\bx^{M_2} \sim p(\bx^{M_2})$ to steer the generation process to a desired content, i.e., aligned the intermediate generative content with the input condition. 
To achieve this goal, we devise a diffusion latent aligner that guides the intermediate noisy latent towards a target direction to the content that the condition depicted during the denoising process.
Formally, given a sequence of latent variables ${\bz_t, \bz_{t-1}, ..., \bz_0}$ from an LDM, the diffusion latent aligner $\mathcal{A}$ takes the corresponding latent $\bz_t$ at arbitrary timestep $t$ alongside the guided condition $\bx^{M_2}$, and produce a modified latent $\hat{\bz_t}$ which has better alignment with the condition.

\begin{align}
    \hat{\bz}_{t}^{M_1} = \mathcal{A}(\bz_{t}^{M1}, \bx^{M_2}).
\end{align}
For joint visual-audio generation, the aligner should simultaneously obtain information from the two modalities and provide guidance signals to these latents:
\begin{align}
    (\hat{\bz}_{t}^{M_1}, \hat{\bz}_{t}^{M_2}) = \mathcal{A}(\bz_{t}^{M_1}, \bz_{t}^{M_2}).
\end{align}
After the sequential denoising process, the goal of our aligner is to minimize the $\mathcal{F}(\mathcal{D}(\bz_{0}^{M_1}), \bx^{M_2})$, for unidirectional guidance, and $\mathcal{F}(\mathcal{D}(\bz_{0}^{M_1}), \mathcal{D}(\bz_{0}^{M_2}))$ for synchronized bidirectional guidance,
where $\mathcal{F}$ indicates a distance function to measure the degree of alignment between samples with two modalities. The updatable parameters in this process can be latent variables, embedding vectors, or neural network parameters.

\subsubsection{Multimodal guidance}
To design such a latent aligner stated in Section~\ref{sec:formulation}, we propose a training-free solution that leverages the great capability of a multimodal model trained for representation learning, i.e., ImageBind~\cite{girdhar2023imagebind} to provide rational guidance on the denoising process. 
Specifically, given latent variables $\bz_t$ at each timestep $t$, the predicted $\bz_0$ can be derived from $\bz_t$ and the predicted noise $\hat{\mathbf{\epsilon}}$ via
\begin{equation}
    \Tilde{\bz}_0 = \mathcal{G}(\bz_t) = \frac{1}{\sqrt{\bar\alpha_t}}\bz_t - \sqrt{\frac{1-\bar\alpha_t}{\bar\alpha_t}}\hat{\mathbf{\epsilon}},
\end{equation}
where $\hat{\mathbf{\epsilon}} = \mathbf{\epsilon}_\theta(\bz_t, t)$.
With such a clean prediction, we can leverage the external models that are trained on normal data without retraining them on noisy data like the classifier guidance is needed.
We feed the $\bz_0$ and the guiding condition to the ImageBind model to compute their distance in the ImageBind embedding space.
The obtained distance can then serve as a penalty, which can be used to backpropagate the computation graph and obtain a gradient on the latent variable $\bz_t$:
\begin{align}
    \mathcal{L}(\Tilde{\bz}_0, \bx^{M_2}) = 1 - \mathcal{F}(\mathbf{E}^{M_1}(\Tilde{\bz}_0), \mathbf{E}^{M_2}(\bx^{M_2})).
\end{align}
Then we update the $\bz_t$ via
\begin{align}
    \hat{\bz}_t = {\bz}_t - \lambda_1 \nabla_{\bz_t} \mathcal{L}(\mathcal{D}(\Tilde{\bz}_0), \bx^{M_2}),
\end{align}
where $\lambda_1$ serves as the learning rate of each optimization step.
In this way, we alter the sampling trajectory at each timestep through our multimodal guidance signal to achieve both audio-to-visual and visual-to-audio.
This procedure only costs a small amount of extra sampling time, without any additional datasets and expensive network training.

\begin{algorithm}[t]
    \caption{Multimodal guidance for joint-VA generation}
    \label{algo:progressive}
    \begin{algorithmic}[1]
        \REQUIRE Learning rate $\lambda_1$, $\lambda_2$, optimization steps $N$, warmup steps $K$, prompt $p$
        \STATE $\mathbf{y} = \textsc{Emb}(p)$
        \FOR{$t=T$ to 0}
            \STATE $\mathbf{z}_{t}^v \xleftarrow{} \textsc{Denoise}(\mathbf{z}_{t+1}^v, \mathbf{y}) $
            \STATE $\mathbf{z}_{t}^a \xleftarrow{} \textsc{Denoise}(\mathbf{z}_{t+1}^a, \mathbf{y}) $
            \IF{$t < K$}
                \FOR{$n=0$ to $N$}       
                    \STATE $\Tilde{\mathbf{z}}_0^v =\frac{1}{\sqrt{\bar{\alpha}_t^v}}\left(\mathbf{z}_t^v-\sqrt{1-\bar{\alpha}_t^v} \mathbf{\epsilon}_t^v\right)$
                    \STATE $\Tilde{\mathbf{z}}_0^a =\frac{1}{\sqrt{\bar{\alpha}_t^a}}\left(\mathbf{z}_t^v-\sqrt{1-\bar{\alpha}_t^a} \mathbf{\epsilon}_t^a\right)$
                    \STATE $\mathbf{e}_a, \mathbf{e}_v, \mathbf{e}_p = \textsc{ImageBind}(\mathbf{z}_0^a, \mathbf{z}_0^v, P)$
                    \STATE $\mathcal{L_{\text{joint-va}}} = \mathcal{F}(\mathbf{e}_v, \mathbf{e}_p) + \mathcal{F}(\mathbf{e}_v, \mathbf{e}_a) + \mathcal{F}(\mathbf{e}_a, \mathbf{e}_p)$
                    \STATE $\hat{\mathbf{z}}_{t}^v = \mathbf{z}_{t}^v - \lambda_1 \nabla_{\mathbf{z}_{t}^v} \mathcal{L}$
                    \STATE $\hat{\mathbf{z}}_{t}^a = \mathbf{z}_{t}^a - \lambda_1 \nabla_{\mathbf{z}_{t}^a} \mathcal{L}$
                    \STATE $\hat{\mathbf{y}} = \mathbf{y} - \lambda_2 \nabla_{\mathbf{y}} \mathcal{L}$
                \ENDFOR
            \ENDIF            
        \ENDFOR
        \RETURN $\mathbf{z}_0^v, \mathbf{z}_0^a$
    \end{algorithmic}
\end{algorithm}

\subsubsection{Dual/Triangle loss function}
We observed that audio often lacks enough semantic information such as some audio is pure background music, while the paired video contains rich semantic information such as multiple objects and environment sound.
Using this type of condition to guide visual generation is not enough and may provide useless guidance information.
To solve this, we incorporate another modality, e,g., text, to provide a comprehensive measurement as
\begin{equation}
    \mathcal{L}_{a2v} = \mathcal{F}(\mathbf{e}_{v}, \mathbf{e_{a}}) + \mathcal{F}(\mathbf{e}_{v}, \mathbf{e_p}).
\end{equation}
The $\mathbf{e}_{v}$, $\mathbf{e}_{a}$ and $\mathbf{e}_{p}$ are the corresponding embeddings in the multimodal space of ImageBind. The $\mathcal{F}$ represents the distance function between two embedding vectors which is one minus cosine similarity between them.
Similarly, the loss for V2A can be written as
\begin{equation}
    \mathcal{L}_{v2a} = \mathcal{F}(\mathbf{e}_{a}, \mathbf{e_{v}}) + \mathcal{F}(\mathbf{e}_{a}, \mathbf{e_p}).
\end{equation}
For visual-audio joint generation, the loss turns into a triangle:
\begin{equation}
    \mathcal{L_{\text{joint-va}}} = \mathcal{F}(\mathbf{e}_v, \mathbf{e}_p) + \mathcal{F}(\mathbf{e}_v, \mathbf{e}_a) + \mathcal{F}(\mathbf{e}_a, \mathbf{e}_p).
\end{equation}

The text can be input by the user to provide a user-guided interactive system or can be extracted via audio captioning models.
As stated before, the audio tends to present incomplete semantic information. Thus, the extracted caption should be worse than that. However, we empirically find that our approach helps to correct these semantic errors, and improves the semantic alignment.
\subsubsection{Guided prompt tuning}
Using the aforementioned multimodal latent guidance, we successfully achieved good generation quality and better content alignment on visual-to-audio generation.
However, we observed that when applying this approach to audio-to-visual generation, the guidance has a neglectable effect.
Meanwhile, when leveraging the audio to generate corresponding audios, the generated video becomes less temporal consistent due to the gradient of each frame having no ensure of temporal coherence.
Therefore, to overcome this issue, we further propose guided prompt tuning by optimizing the input text embedding vector of the generative model, which is formulated as
\begin{align}
    \hat{\by} = \by - \lambda_2 \nabla_{\by} \mathcal{L}.
\end{align}
The $\lambda_2$ indicates the learning rate for the prompt embedding.
Specifically, we detach the prompt text embedding at the beginning of predicting the noise and retain a computational graph from the text embedding to the calculation of multimodal loss. 
Then we backpropagate the computational graph to obtain the gradient of the prompt embedding \textit{w.r.t.} the multimodal loss.
The updated embedding is shared across all timesteps to provide consistent semantic guidance information.

\begin{table*}[t]
    \centering
    \setlength\tabcolsep{5pt} 
    \begin{tabular}{cccccc}
        Task & Method & \multicolumn{4}{c}{Metric}  \\
        \toprule
        \multirow{4}{*}{V2A} 
        &  & \cellcolor[rgb]{1, 1, 0.9}{KL}$\downarrow$ & \cellcolor[rgb]{1, 1, 0.9}{ISc}$\uparrow$ & \cellcolor[rgb]{1, 1, 0.9}{FD}$\downarrow$ & \cellcolor[rgb]{1, 1, 0.9}{FAD}$\downarrow$ \\
        \cline{3-6}
        & SpecVQGAN~\cite{iashin2021specvqgan} & 3.290 & 5.108 & 37.269 & 7.736  \\
        & Ours-vanilla & 3.203 & 5.625 & 40.457 & \textbf{6.850} \\
        & Ours & \textbf{2.619} & \textbf{5.831} & \textbf{32.920} & \underline{7.316} \\
        \hline
        \multirow{3}{*}{I2A} 
        &  & \cellcolor[rgb]{1, 1, 0.9}{KL}$\downarrow$ & \cellcolor[rgb]{1, 1, 0.9}{ISc}$\uparrow$ & \cellcolor[rgb]{1, 1, 0.9}{FD}$\downarrow$ & \cellcolor[rgb]{1, 1, 0.9}{FAD}$\downarrow$ \\
        \cline{3-6}
        & Im2Wav~\cite{sheffer2023im2wav} & 2.612 & \textbf{7.055} & \textbf{19.627} & 7.576 \\
        & Ours-vanilla & 3.115 & 4.986 & 33.049 & 7.364 \\
        & Ours & \textbf{2.691} & 6.149 & 20.958 & \textbf{6.869} \\
        \hline
        \multirow{3}{*}{A2V} 
        &  & \cellcolor[rgb]{1, 1, 0.9}{FVD}$\downarrow$ & \cellcolor[rgb]{1, 1, 0.9}{KVD}$\downarrow$ & \cellcolor[rgb]{1, 1, 0.9}{AV-align}$\uparrow$ & - \\
        \cline{3-6}
        & TempoToken~\cite{yariv2023tempotokens} & 1866.285  & 389.096 & 0.423 & - \\
        & Ours-vanilla & 417.398 & 36.262 & 0.518 & - \\
        &  Ours & \textbf{402.385} & \textbf{34.764} & \textbf{0.522} & - \\
        \hline
        \multirow{6}{*}{Joint VA Generation} 
        &  & \cellcolor[rgb]{1, 1, 0.9}{FVD}$\downarrow$ & \cellcolor[rgb]{1, 1, 0.9}{KVD}$\downarrow$ & \cellcolor[rgb]{1, 1, 0.9}{FAD}$\downarrow$ \\
        \cline{3-6}
        & Landscape: MM~\cite{ruan2023mm} & \textbf{1141.009} & \textbf{135.368} & 7.752 & - \\
        & Landscape: MM~\cite{ruan2023mm} + Ours & 1174.856 & 135.422 & \textbf{6.463} & - \\
        \hhline{~:-----}
        &  & \cellcolor[rgb]{1, 1, 0.9}{$\text{AV-align}_{bind}$}$\uparrow$ & \cellcolor[rgb]{1, 1, 0.9}{$\text{VT-align}_{bind}$} $\uparrow$ & \cellcolor[rgb]{1, 1, 0.9}{$\text{AT-align}_{bind}$} $\uparrow$ & \cellcolor[rgb]{1, 1, 0.9}{AV-align} $\uparrow$ \\
        \cline{3-6}
        & Open-domain: MM\cite{ruan2023mm} & N/A & N/A & N/A & N/A \\
        & Open-domain: Ours-vanilla & 0.074 & 0.322 & 0.081 & 0.226 \\
        & Open-domain: Ours & \textbf{0.096} & \textbf{0.324} & \textbf{0.138} & \textbf{0.283} \\
        \bottomrule
    \end{tabular}
    \caption{Quantitative comparison with baselines on four tasks.}
    \label{tab:main}
\end{table*}

\begin{figure*}[t]
    \centering
    \includegraphics[width=1.0\linewidth]{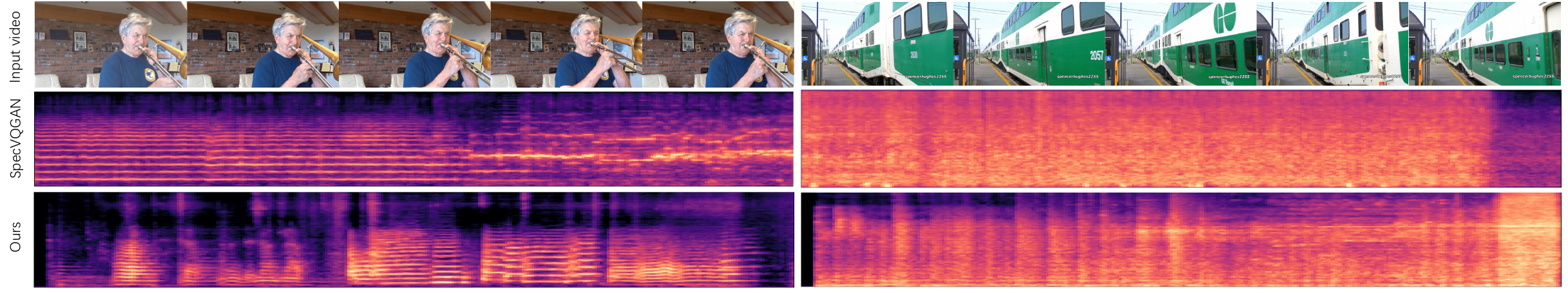}
    \caption{Compared with baseline on the video-to-audio generation task. SpecVQGAN fails to generate realistic and aligned audio with the input video. Our method can produce aligned audio with the input video rhythm.}
    \label{fig:v2a}
\end{figure*}

\begin{figure*}[t]
    \centering
    \includegraphics[width=1.0\linewidth]{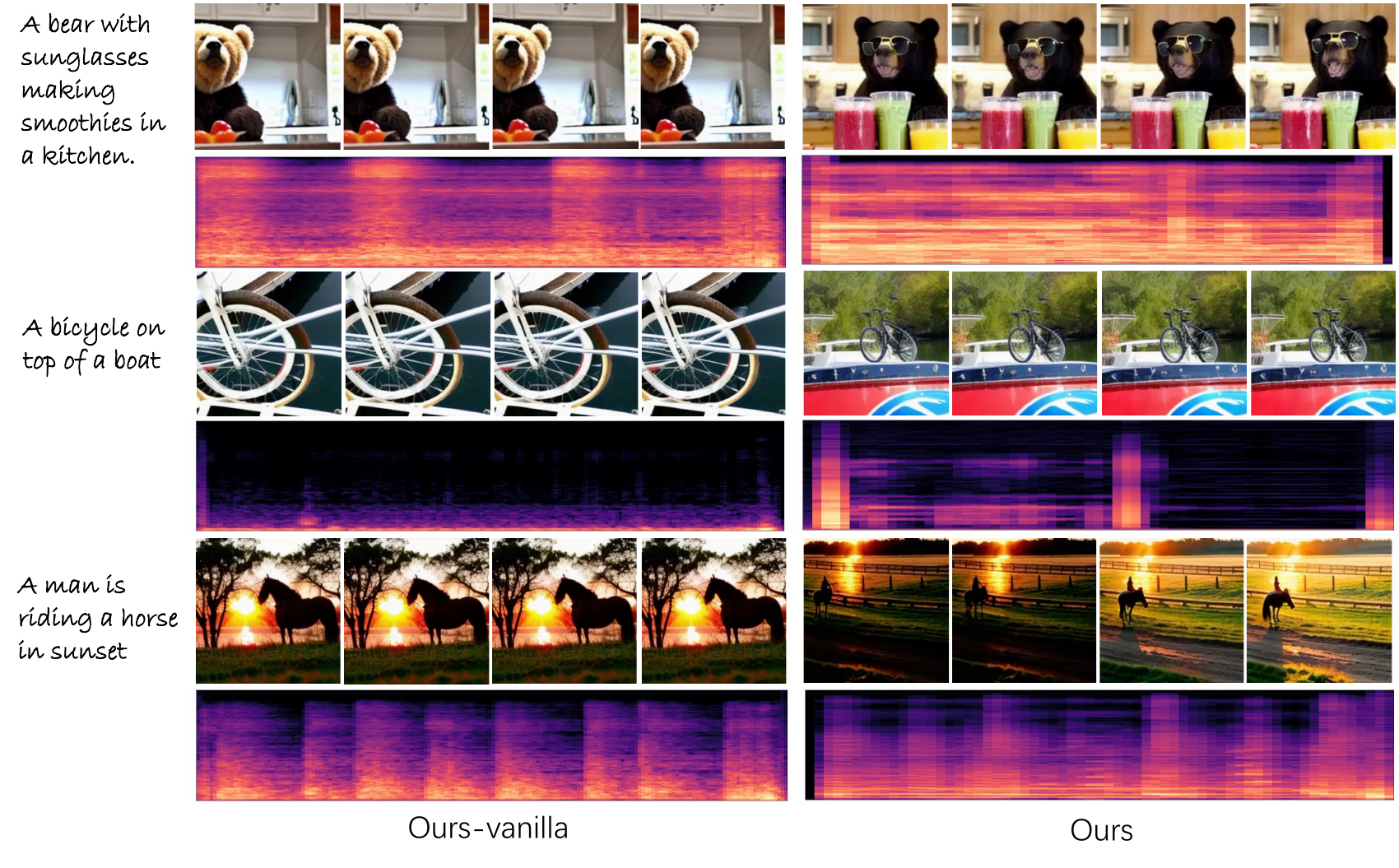}
    \vspace{-8.5mm}
    \caption{Compared with baseline on the joint video-and-audio generation task. Our method can produce better text-aligned visual content than the vanilla model. Besides, our generated audio is also of better quality and better alignment with the generated videos.}
    \label{fig:joint_ablation}
    \vspace{-3mm}
\end{figure*}

\begin{figure*}[t]
    \centering
    \includegraphics[width=1.0\linewidth]{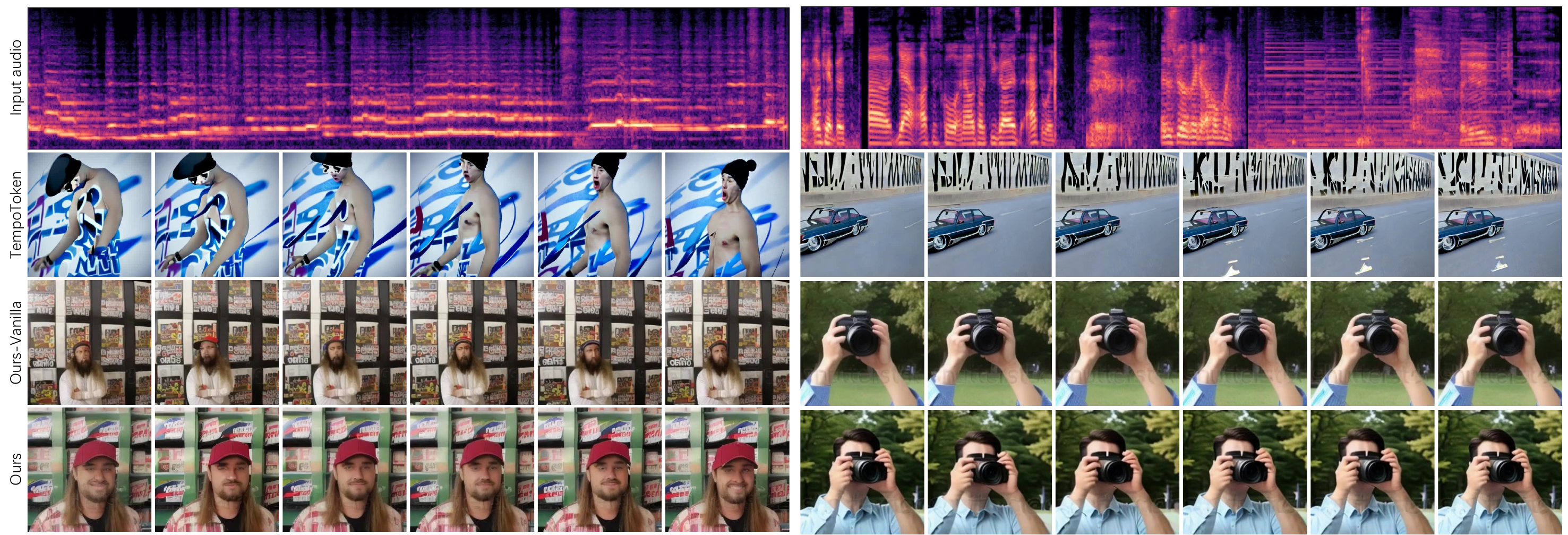}
    \vspace{-4mm}
    \caption{Compared with baseline on the audio-to-video task. Given the input audio, the generated videos by TempoToken are not aligned with the input audio and the generation with poor visual quality. Our method can produce visually much better and semantically aligned content with the input condition. }    
    \label{fig:a2v}
    \vspace{-3mm}
\end{figure*}


\section{Experiments}

\subsection{Experimental Setup}
\textbf{Dataset} We utilize the VGGSound dataset~\cite{chen2020vggsound} and Landscape dataset~\cite{ruan2023mm} for evaluation on video-to-audio, audio-to-video, and audio-video joint generation task. Since our method is an optimization-based solution, there is no need to utilize the entire dataset for evaluation. Instead, we randomly sample 3k video-audio pairs from the VGGSound dataset for video-to-audio generation, 3k pairs for audio-to-video generation, and 3k pairs for image-to-audio generation respectively. We extract the key frame from each video for the image-to-audio generation task.
We also randomly sample 200 video-audio pairs from the Landscape dataset for video-audio joint generation. 

\noindent\textbf{Implementation details}
We utilize the pretrained AudioLDM~\cite{liu2023audioldm} for video-to-audio and image-to-audio generation, the AnimateDiff~\cite{guo2023animatediff} for audio-to-video generation. We use both the pre-trained AudioLDM and AnimateDiff for the joint audio-video generation. 
We set the denoising step to 30 for video-to-audio generation, 25 for audio-to-video generation, and 25 for audio-video joint generation, respectively. 
We use the learning rate 0.1 for guiding the AudioLDM denoising and 0.01 for guiding the AnimateDiff denoising, which applies to all the tasks. We fixed the random seed of the optimization process for fair comparisons. 
All the experiments are conducted on NVIDIA Geforce RTX 3090 GPUs. 
 
\subsection{Baselines}
\textbf{Video-to-Audio} We choose SpecVQGAN~\cite{iashin2021specvqgan} as the baseline of Video-to-Audio generation task. We used the pre-trained model, which was trained using ResNet-50 with 5 features on VGGSound~\cite{iashin2021specvqgan} as our inference model and compared our method with SpecVQGAN on 3k VGGSound sample datasets.

\noindent\textbf{Image-to-Audio} We choose Im2Wav as the baseline of the Image-to-Audio generation task and used the pre-trained model provided by the authors~\cite{sheffer2023im2wav}, test on 3k Paprika style transferred VGGSound samples transferred by AnimeGANv2~\cite{AnimeGANv2}.

\noindent\textbf{Audio-to-Video} We choose TempoTokens as the baseline of the Audio-to-Video generation task and used the pre-trained model provided by the authors~\cite{yariv2023tempotokens}, test on 3k VGGSound samples.

\noindent\textbf{Joint video and audio generation} As MM-Diffusion~\cite{ruan2023mm} is the state-of-the-art of unconditional video and audio joint generation task, We choose it as the baseline of unconditional video and audio joint generation task in the limit Landscape domain with 200 Landscape samples using the model pre-trained on Landscape datasets. On the open domain, we compare our Ours-with-guidance model with the Ours-vanilla model, as, to the best of our knowledge, there is no established baseline for this task.

\noindent\textbf{Ours-Vanilla} We design several vanilla models of our tasks with the combination of existing tools. For the video-to-audio task, we extract the key frame~\cite{katna} and use a pre-trained image caption model~\cite{qwen} to obtain the caption for the video. We then use the extracted caption to generate audio with the AudioLDM model. For the audio-to-video task, we use an audio caption model and feed the extracted caption to the AnimateDiff to generate the videos for the input audio. For the joint audio and video generation task, we directly take the test prompt as input to the AudioLDM model and AnimateDiff model to compose the joint generation results.

\subsection{Visual-to-Audio Generation}
\begin{figure}[t]
    \centering
    \includegraphics[width=1.0\linewidth]{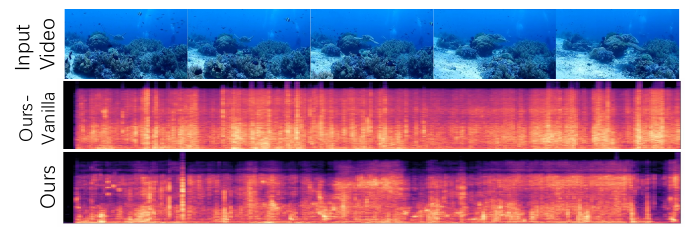}
    \vspace{-7mm}
    \caption{Compared with our vanilla model on the video-to-audio generation task. Our method can significantly reduce the background and irrelevant sound and thus achieve better audio quality, which is also reflected in Tab.~\ref{tab:main}.}
    \label{fig:v2a_ablation}
    \vspace{-3mm}    
\end{figure}

Visual-to-audio generation includes video-to-audio generation and image-to-audio generation tasks. The image-to-audio generation requires audio-visual alignment from the semantic level, whereas temporal alignment is additionally needed for video-to-audio generation. Moreover, the generated audio also needs to be high-fidelity. 
To quantitatively evaluate our performance on these aspects, we utilize the MKL metric~\cite{iashin2021specvqgan} for audio-video relevance, Inception score (ISc), Frechet distance (FD), and Frechet audio distance (FAD) for audio fidelity evaluation. From Tab.~\ref{tab:main}, we can see that even though our method is training-free, we can still outperform the baseline which requires large-scale training on audio-video pairs. When compared with the text-to-audio baseline, we could see that our method consistently improves the audio-video relevance and the audio generation quality. When compared with our vanilla baseline, we find our method can significantly improve the audio quality, especially by reducing irrelevant sound and background noise, as shown in Fig.~\ref{fig:v2a_ablation}.

\subsection{Audio-to-Video Generation}
Audio-to-video generation requires the generated videos to be high-quality, as well as semantically and temporally aligned with the input audio. To quantitatively evaluate the visual quality of the generated videos, we adopt the Frechet Video Distance (FVD) and Kernel Video Distance (KVD)~\cite{unterthiner2018towards} as the metrics. We also use the audio-video alignment (AV-align)~\cite{yariv2023tempotokens} metric to measure the alignment of the generated video and the input audio. We show our quantitative results in Tab.~\ref{tab:main}. We observe that our training-free method can outperform the training-based baseline in terms of both semantic alignment and video quality. Besides, compared with the text-to-video method, our method can achieve better audio-video alignment while maintaining a comparable visual quality performance. We show our qualitative results in Fig.~\ref{fig:a2v}. We observe that TempoToken struggles with visual quality and audio-visual alignment, and thus the generated videos are not relevant to the input audio and the generated quality is relatively poor. Although the text-to-video method can achieve good performance on the visual quality of the generated videos, it struggles to accurately align with the input audio content. Our training-free method, utilizing a shared audio-visual representation space, can achieve a good tradeoff between visual quality and audio-visual alignment.

\begin{figure}[t]
    \centering
    \includegraphics[width=1.0\linewidth]{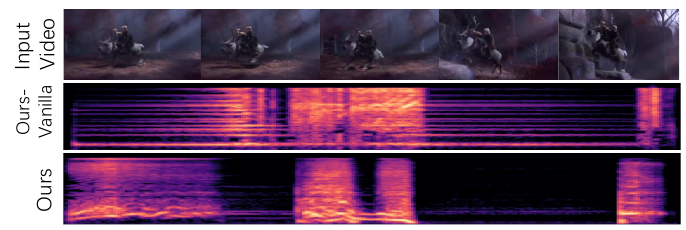}
    \vspace{-5mm}
    \caption{We visualize the effect of our guided prompt tuning. The automatic caption generated is ``frozen 2 - screenshot", which fails to capture the meaningful visual content, and thus, the text-to-audio method fails to produce meaningful sounds. 
    Our prompt tuning can inspect the visual information to complement the semantic information to generate meaningful sounds.
    }
    \label{fig:v2a_ablation_2}
    \vspace{-5mm}
\end{figure}

\subsection{Joint Video and Audio Generation}
The practical joint video and audio generation task should take the text as the input, produce high-fidelity videos and audio, maintain the audio-video alignment, and maintain the text-audio and text-video relevance. Specifically, we adopt the FVD for video quality, FAD for audio quality, AV-align~\cite{yariv2023tempotokens} for audio-video relevance, TA-align for text-audio alignment, and the TV-align for text-video alignment. Our quantitative evaluation is shown in Tab.~\ref{tab:main}. Our latent aligner can be plugged into existing unconditional audio-video joint generation framework MM-Diffusion~\cite{ruan2023mm}. The results show that compared with the original MM-Diffusion, our latent aligner can boost the audio generation quality when maintaining the video generation performance. We also verify our method of text-conditioned joint video and audio generation. We bridge the video diffusion model AnimateDiff~\cite{guo2023animatediff} and audio diffusion model AudioLDM~\cite{liu2023audioldm} with our diffusion latent aligner. We randomly collect 100 prompts from the web to condition our generation.
Compared with separate text-to-video and text-to-audio models, our aligner can improve text-video alignment, text-audio alignment, and video-audio alignment. We show the qualitative comparison in Fig.~\ref{fig:joint_ablation}. More qualitative results can be found in the Supplementary.

\subsection{Limitations}
Our performance is limited by the generation capability of the adopted foundation generation models, i.e., AudioLDM and AnimateDiff. 
For example, for our A2V and Joint VA tasks built on AnimateDiff, the visual quality, complex concept composition, and complex motion could be improved in the future.
Notably, the flexibility of our method allows for adopting more powerful generative models in the future to further improve the performance. 
\section{Conclusion}
We propose an optimization-based method for the open-domain audio and visual generation task. Our method can enable video-to-audio generation, audio-to-video generation, video-audio joint generation, image-to-audio generation, and audio-to-image generation tasks. Instead of training giant models from scratch, we utilize the a shared multimodality embedding space provided by ImageBind to bridge the pre-trained visual generation and audio generation diffusion models. Through extensive experiments on several evaluation datasets, we show the advantages of our method, especially in terms of improving the audio generation fidelity and audio-visual alignment. 

\paragraph{Acknowlegement} This project was supported by the National Key R\&D Program of China under grant number 2022ZD0161501. 

{
    \small
    \bibliographystyle{ieeenat_fullname}
    \bibliography{main}
}


\end{document}